\pdfoutput=1

\documentclass[letterpaper, 10pt, conference]{ieeeconf}

\IEEEoverridecommandlockouts

\usepackage{xcolor}
\usepackage{colortbl}
\definecolor{liftpotbg}{rgb}{0.950, 0.842, 0.779}
\definecolor{closemarkerbg}{rgb}{0.766, 0.815, 0.880}
\definecolor{platehandoverbg}{rgb}{0.830, 0.747, 0.826}
\usepackage{amsmath} %
\usepackage{amssymb}  %
\usepackage{multirow}
\usepackage{graphicx}
\usepackage{placeins}
\usepackage{stfloats}
\usepackage{cite}

\usepackage[hidelinks]{hyperref}
\usepackage[capitalize,noabbrev]{cleveref}
\usepackage{caption}
\usepackage{xspace}
\newcommand{\longours}{Unified Visuomotor Target (UVT) Training}
\newcommand{\shortours}{UVT\xspace}
\newcommand{\website}{\href{https://unified-visuomotor-targets.github.io/}{\texttt{unified-visuomotor-targets.github.io}}\xspace}
\newcommand{\webpage}{\href{https://unified-visuomotor-targets.github.io/}{webpage}\xspace}

\title{\LARGE \bf
Unified Visuomotor Targets: \\
Supervising VLAs Beyond Physical Actions
}

\author{Zhenyang Feng$^{1}$ and Unnat Jain$^{1}$%
\thanks{$^{1}$Zhenyang Feng and Unnat Jain are with the Department of Computer Science, University of California, Irvine.}%
}

\begin{document}

\maketitle
\thispagestyle{empty}
\pagestyle{empty}

\begin{abstract}
VLA models are trained to predict robot actions from visual and
language observations.
This is a natural choice, but it creates a mismatch: VLMs encode
rich, high-level representations of scenes and goals, while robot
actions are low-level signals with limited task structure.
We ask whether changing \emph{what} the policy is trained to predict,
rather than how it is architecturally designed, can yield better and
more efficiently trained policies.
We propose \shortours~(Unified Visuomotor Target), a unified latent
prediction target that jointly encodes motor control and visual scene
transition information, requiring no architectural changes and no additional data.
Applied to two representative VLA systems across simulation benchmarks
and real bimanual manipulation tasks, \shortours~improves training
efficiency, final task performance, and policy robustness, with
particularly strong gains under limited training budgets and
challenging environmental conditions.
Rollout videos and additional qualitative results are available at
our project webpage: \website.
\end{abstract}

\section{Introduction}
\label{sec:intro}

Vision-Language-Action (VLA) models adapt pretrained vision-language
models (VLMs) as robot policy backbones, mapping image observations and
language instructions to robot actions.
Early systems extended the VLM vocabulary to predict discretized action
tokens, treating binned joint values or frequency-domain codes as
language tokens~\cite{rt2,pi0fast,vqvla}.
Treating actions as tokens aligns with autoregressive VLM pretraining,
but discretization limits motion precision.
Predicting raw continuous actions via direct regression or flow matching
recovers this precision, but VLMs
are not pretrained to produce continuous actuator signals.
Recent work has addressed this by specializing the architecture:
MLP heads on top of VLM predictions~\cite{openvla_oft}, action expert
transformers conditioned on VLM key-value features~\cite{pi0,pi},
and cross-attention modules adapted for robotic control~\cite{adapter}.
Adding these modules can degrade the VLM's pretrained representations,
motivating separate techniques to preserve them~\cite{ki,anchoralign}.
\emph{All of these ask how to adapt the architecture to the target,
not what the target should be.}

The targets remain the same: raw robot actions, first quantized, now continuous.
VLMs are trained to predict task-relevant semantic outputs from image
and language inputs: which objects are present, their spatial
relations, what the goal state is.
Robot joint angles and Cartesian poses have none of this structure;
the same wrist motion or gripper correction recurs across tasks with
entirely different goals.
Training a semantic predictor to output kinematic signals slows
convergence and limits robustness.
We ask: \textbf{can a prediction target better aligned with VLM
representations improve VLA training without architectural changes?}

Latent Action Models (LAMs)~\cite{lapa,univla,villax,mvplam,routray2025vipra} encode
what robot actions lack.
Trained as inverse-dynamics models on image pairs, LAMs compress the
visual state transition between two frames into a compact discrete code
that captures task-relevant scene change, not embodiment-specific motor
commands.
These models have been used primarily for large-scale robotics
pretraining, where dynamics codes bootstrap generalist policies from
video and cross-embodiment data.
Their use as a fine-tuning supervision signal has not been explored.

We propose \shortours~(Unified Visuomotor Target), a compact latent
supervision target for VLA fine-tuning that jointly encodes robot actions
and visual state-transition codes from a pretrained LAM.
As shown in Fig.~\ref{fig:pipeline}, a lightweight multimodal VAE, trained
once on the demonstration dataset, fuses each action chunk with its
corresponding LAM code into a 32-dimensional latent.
During fine-tuning, the VLA is supervised on this latent rather than raw
actions; a trainable decoder recovers an executable action chunk at inference.
Because the target captures both motor control and observed scene changes as discrete codes,
it is better matched to what VLMs encode from image and language inputs
than raw joint signals.
The approach requires no architectural changes, no additional data, and
no modification to the training schedule.

We evaluate \shortours~on VLA-Adapter~\cite{adapter} (with regression head)
and $\pi_{0.5}$~\cite{pi} (with flow-matching head) on LIBERO~\cite{libero},
the harder LIBERO-Plus benchmark~\cite{liberoplus}, and three real-world
bimanual manipulation tasks.
On both VLA architectures, \shortours improves convergence speed,
task performance, and policy robustness.
Under LIBERO-Plus, \shortours reaches 81.5\% on Spatial tasks at 10k
steps versus 34.0\% for the baseline.
On real robot tasks, \shortours substantially improves grasping
reliability: on \textit{Close Marker}, the baseline achieves 0\%
success across 50 trials while \shortours reaches 38\%; on
\textit{Plate Handover}, success improves from 18\% to 40\%.

Our contributions are threefold: (1) We introduce \shortours, a prediction target that fuses robot
    actions with visual dynamics codes from a pretrained LAM, requiring
    no architectural changes and no additional data. (2) We show that training VLAs on \shortours rather than raw
    actions improves training efficiency by providing a supervision
    signal more aligned with what VLMs encode. (3) We demonstrate improvements in convergence speed, task
    performance, and robustness across regression and flow-based VLA
    policies on LIBERO, LIBERO-Plus, and three real bimanual
    manipulation tasks.

\section{Related Work}
\label{sec:related}

\noindent\textbf{Vision-Language-Action Models.}
Vision–Language–Action (VLA) models repurpose pretrained Vision–Language Models (VLMs) as multimodal backbones for robotic control, leveraging their strong visual–textual reasoning capabilities~\cite{rt2,openvla,pi}. Existing VLA approaches primarily differ in how actions are decoded from VLM representations. 

The first line of work towards this direction extends the VLM vocabulary to directly predict discretized action tokens that are subsequently mapped to continuous control~\cite{openvla,vqvla,pi0fast}. More recent models instead attach a dedicated policy head on top of VLM hidden states to produce continuous actions: diffusion or flow-matching based heads generate actions through iterative denoising~\cite{chi2023diffusionpolicy,pi0,pi}, while regression-style heads (e.g., MLPs or transformer decoders) predict continuous action chunks in parallel~\cite{adapter,roboflamingo}. Despite architectural differences, these paradigms typically supervise the model using low-level robot actions.

From a training perspective, most general-purpose policies follow a two-stage paradigm consisting of large-scale robotics pretraining followed by task-specific fine-tuning. Pretraining is often performed on aggregated multi-robot datasets~\cite{oxe,octo} or image/video datasets~\cite{bahl2023vrb,dasari2023data4robotics,molmoact,patel2026rigvid} requiring substantial computational resources to learn cross-task and cross-embodiment motion priors. The pretrained model is then adapted to a specific environment using a smaller demonstration dataset in the target domain.

Recent work has also explored bypassing large-scale robotics pretraining altogether. Methods such as VLA-Adapter~\cite{adapter} demonstrate that pretrained VLMs can be directly adapted into robotic policies via lightweight policy heads and fine-tuning, significantly reducing training cost. Complementary work studies how to fine-tune VLAs without overwriting pretrained VLM representations~\cite{ki,anchoralign}. These results suggest that efficient adaptation matters for scalable VLA deployment, even when large-scale pretraining helps.

\medskip
\noindent\textbf{Latent Action Learning.}
Latent Action Models (LAMs) aim to abstract action dynamics beyond explicit robot actions, enabling cross-embodiment and cross-domain generalization. Building on latent-action models from video such as Genie~\cite{genie}, LAPA~\cite{lapa} introduces a Latent Action Model (LAM) trained with a variational objective to encode frame-to-frame dynamics into a compact discrete code space. By modeling visual transitions rather than robot actions, LAMs capture higher-level visual transition that can be shared across different robot embodiments and even human video data.

Subsequent works extend this idea in various directions. UniVLA~\cite{univla} proposes task-centric dynamics codes to improve generalization across diverse environments. ConLA~\cite{conla} and MVP-LAM~\cite{mvplam} explore contrastive and cross-view reconstruction objectives to enhance the consistency and expressiveness of dynamics codes. VQ-VLA~\cite{vqvla} further investigates vector-quantized action tokenizers within VLA training. All these methods aim to demonstrate that structured dynamics code representations can provide more abstract and transferable motion priors compared to robot actions.

So far, most of these approaches have been studied primarily in the context of large-scale robotics pretraining, where dynamics codes are used to bootstrap generalist policies. In contrast, our work focuses on a different setting: we use pretrained LAMs as auxiliary regularizers during downstream VLA fine-tuning. Rather than replacing action supervision, we integrate dynamics code signals with robot actions to construct a structured prediction target that improves optimization efficiency across both flow-matching and regression-style VLA policies.

\section{\longours}
\label{sec:approach}

\begin{figure*}[!ht]
\centering
\includegraphics[width=0.95\textwidth]{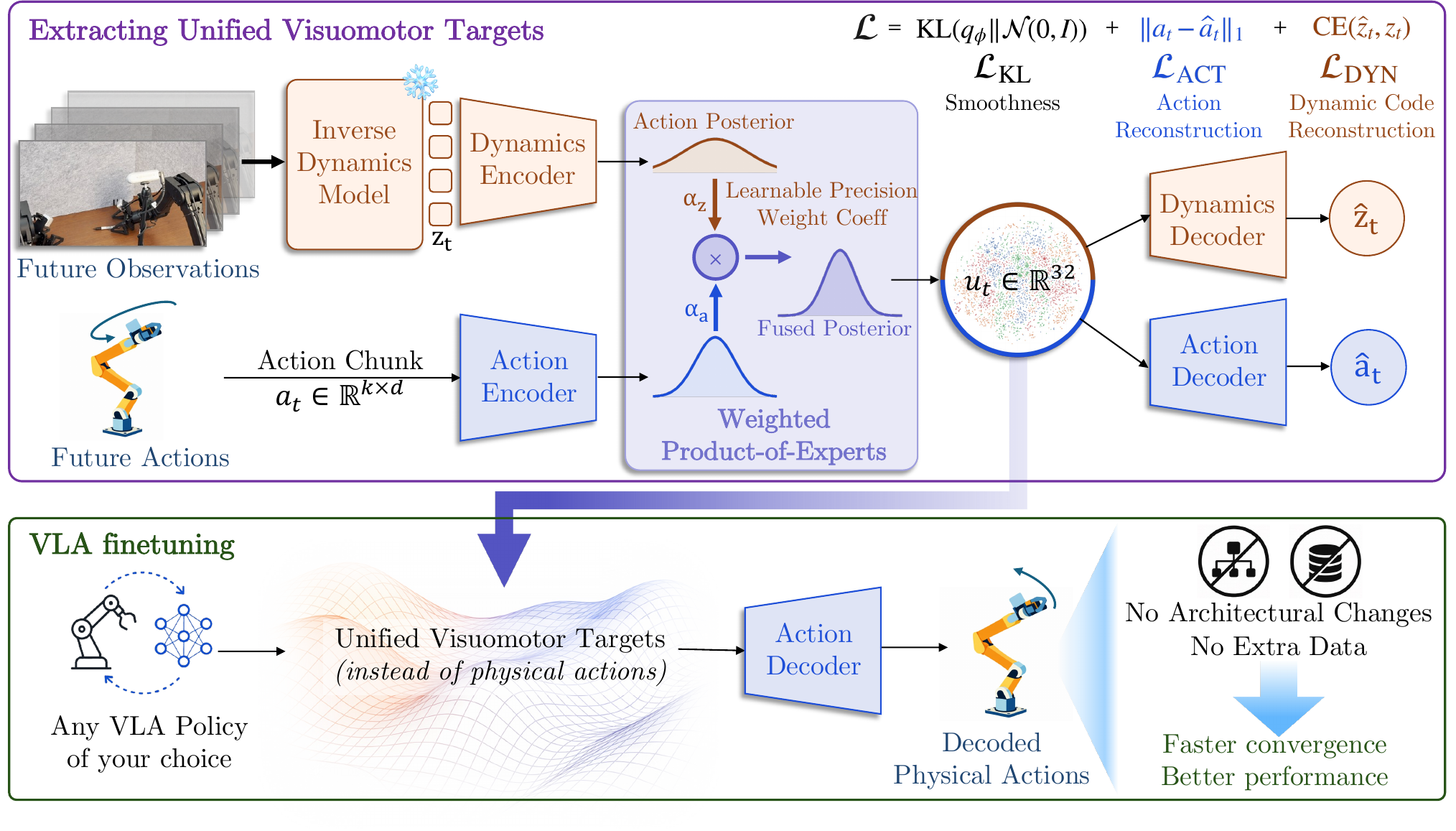}
    \caption{\textbf{Unified Visuomotor Target (\shortours) Overview.}
\emph{Top (Extracting \shortours):} a frozen pretrained inverse dynamics model extracts discrete dynamics codes $z_t$ from future observations, while ground-truth action chunks $a_t \in \mathbb{R}^{k \times d}$ ($k$: chunk size, $d$: action dimension) provide continuous motor supervision. Two trainable encoders map $a_t$ and $z_t$ to Gaussian posteriors, which are fused via a weighted Product-of-Experts~\cite{hinton2002poe} with learnable precision weights $\alpha_a, \alpha_z$ to produce the unified visuomotor target $u_t \in \mathbb{R}^{32}$. The target is decoded to reconstruct both robot actions ($\mathcal{L}_{\text{ACT}}$) and dynamics codes ($\mathcal{L}_{\text{DYN}}$), while a KL divergence loss $\mathcal{L}_{\text{KL}}$ regularizes $u_t$ toward a standard normal prior, encouraging a well-conditioned prediction space.
\emph{Bottom (VLA finetuning):} the VLA policy is supervised to predict $u_t$, and a trainable action decoder maps the predicted target back to executable actions, enabling faster convergence and better performance with no architectural changes and no extra data.}
\label{fig:pipeline}
\end{figure*}

\subsection{Preliminaries}
\label{sec:preliminaries}

We consider a robot manipulation setting where demonstrations consist of
multimodal observations and robot actions.
Let $\mathcal{D} = \{(x_t, a_t)\}$ denote a demonstration dataset.
Here, $x_t$ includes visual observations and language instructions at time $t$,
and $a_t \in \mathbb{R}^{k \times d}$ denotes a future ground-truth action chunk the robot should take to complete the task, where $k$ is the number of action steps each forward pass produces and
$d$ is the action dimensionality (e.g., 7D Cartesian pose with gripper).

An action chunk $a_t$ spans the interval $[t, t+k]$ and represents
a short-horizon trajectory of continuous robot actions.
The action-chunk formulation follows~\cite{zhao2023act}.
Modern Vision-Language-Action (VLA) models are typically fine-tuned
to predict such action chunks directly from multimodal inputs \cite{adapter, pi, univla, openvla, octo}:
\[
\min_\theta \;
\mathbb{E}_{(x_t,a_t)\sim\mathcal{D}}
\left[
-\log p_\theta(a_t | x_t)
\right].
\]
This is the standard imitation learning objective used during
VLA fine-tuning.

The pretrained LAM operates purely on visual observations.
Let $v_t$ denote the visual observation component of $x_t$.
Given two images $v_t$ and $v_{t+k}$,
the LAM encodes their 2D visual state transition into a discrete dynamics code
\[
z_t = \mathrm{LAM}(v_t, v_{t+k}),
\]
where $z_t$ captures image-level dynamics across the action window.
Importantly, $z_t$ is derived from visual state transitions rather than
robot actions, and therefore contains no explicit motor grounding.

In what follows, we will construct a unified visuomotor target $u_t$
that integrates robot action $a_t$
with visual transition code $z_t$.

\subsection{Method Overview}
\label{sec:overview}
Our goal is to replace direct robot action supervision
with a Unified Visuomotor Target (UVT),
denoted $u_t$,
that integrates motor control and visual transition information.

As illustrated in Figure~\ref{fig:pipeline},
for each demonstration segment,
we obtain the action chunk $a_t$
and the corresponding discrete dynamics code $z_t$
from a pretrained LAM.
These two signals answer different questions about the same motion.
The action chunk says how the robot moves;
the dynamics code says how the scene changes.
Fusing them into one target gives the VLA a supervision signal that matches
both the motor command and the visual transition it must produce.

We therefore learn a mapping
$u_t = \Phi(a_t, z_t)$,
where $\Phi$ produces a compact UVT representation
that preserves robot action control fidelity while incorporating visual dynamics.

During VLA fine-tuning,
the policy is trained to predict $\hat{u}_t$ from multimodal input $x_t$.
A trainable decoder then translates $\hat{u}_t$ back to executable robot actions.

This framework consists of three components:
(i) a multimodal encoder that constructs $u_t$ from $(a_t, z_t)$,
(ii) a UVT supervision objective,
and (iii) a trainable decoder for action reconstruction, as shown in \cref{fig:pipeline}.
The following subsections describe each component in detail.

\subsection{Unified Visuomotor Target Construction}
\label{sec:uvt_design}

We construct the unified visuomotor target $u_t \in \mathbb{R}^{d_u}$
using a multimodal variational autoencoder (MVAE)~\cite{wu2018multimodal},
as illustrated in Fig.~\ref{fig:pipeline}.
The MVAE maps the robot action $a_t$ and the discrete dynamics code $z_t$
into a shared UVT space and produces a unified visuomotor target $u_t$
that can reconstruct both signals.

\noindent\textbf{Encoders.}
We use two encoders that map each modality into a diagonal Gaussian
in the same UVT space.
The robot action encoder outputs:
\[
(\mu_t^{a}, \log \sigma_t^{2,a})
=
f^{a}(a_t),
\]
and the discrete dynamics encoder outputs:
\[
(\mu_t^{z}, \log \sigma_t^{2,z})
=
f^{z}(z_t).
\]
Thus, each training pair $(a_t, z_t)$ yields two Gaussian predictions over $u$.
Both encoders are trained jointly because the fused UVT (defined next)
must support reconstruction of \emph{both} modalities, which encourages
the two encoders to produce aligned UVT representations.

\noindent\textbf{Fusion.}
We combine the two Gaussian predictions using weighted precision fusion.
Let $\tau = 1/\sigma^2$ denote elementwise precision.
The fused precision and mean are:
\begin{align*}
\tau_t
&=
\alpha_{a}\, \tau_t^{a}
+
\alpha_{z}\, \tau_t^{z},
\\
\mu_t
&=
\frac{
\alpha_{a}\, \tau_t^{a} \odot \mu_t^{a}
+
\alpha_{z}\, \tau_t^{z} \odot \mu_t^{z}
}{
\tau_t
},
\end{align*}
where $\odot$ denotes elementwise multiplication. 
The inverse variance (precision) acts as a weighting factor:
dimensions with smaller predicted variance contribute more strongly to the fused mean.
The learned scalar weights $\alpha_{a}$ and $\alpha_{z}$
provide a global balance between robot action and visual dynamics signals. Finally, we use the fused mean as the unified visuomotor target:
\[
u_t = \mu_t.
\]
The corresponding fused variance is $\sigma_t^{2} = 1/\tau_t$ (elementwise),
which parameterizes the Gaussian used in the KL term below.

\noindent\textbf{Decoders.}
Two decoders reconstruct both modalities from the same UVT:
\[
\hat{a}_t = g^{a}(u_t), \qquad \hat{z}_t = g^{z}(u_t).
\]
This forces $u_t$ to preserve information needed to recover
both the robot action and the discrete dynamics code.

\noindent\textbf{Training objective.}
We train the MVAE with reconstruction losses for both modalities
and a KL regularizer that keeps the UVT distribution close to a standard normal:
\[
\mathcal{L}_{\text{ACT}}
=
\|a_t - \hat{a}_t\|_1,
\qquad
\mathcal{L}_{\text{DYN}}
=
\mathrm{CE}(z_t, \hat{z}_t),
\]
\[
\mathcal{L}_{\text{KL}}
=
D_{\mathrm{KL}}\!\left(
\mathcal{N}(\mu_t, \operatorname{diag}(\sigma_t^2))
\;\|\;
\mathcal{N}(0,I)
\right).
\]
The overall objective is to minimize the term:
\[
\mathcal{L}_{\text{UVT}}
=
\mathcal{L}_{\text{ACT}}
+
\mathcal{L}_{\text{DYN}}
+
\mathcal{L}_{\text{KL}}.
\]

This procedure produces the unified visuomotor target $u_t$
used as the prediction target for VLA fine-tuning.

\subsection{\shortours~for Imitation Learning with VLAs}
\label{app:integration}

The MVAE described in Sec.~\ref{sec:uvt_design}
is trained first to construct the unified visuomotor target $u_t$
from robot actions $a_t$ and discrete dynamics codes $z_t$.
After training, we precompute
$u_t = \Phi(a_t, z_t)$
for all samples in the demonstration dataset $\mathcal{D}$.
VLA imitation learning is then performed in this UVT space,
using the triples $(x_t,u_t,a_t)$ as its data sample.

\noindent\textbf{UVT objective.}
Given multimodal input $x_t$,
the VLA policy predicts
\[
\hat{u}_t = f_\theta(x_t),
\]
and is supervised using
\[
\mathcal{L}_{\text{UVT-align}}
=
\| \hat{u}_t - u_t \|_2^2.
\]

This loss anchors the policy’s internal representation
to the unified visuomotor target space,
encouraging it to predict both robot action
and visual dynamics information.

\noindent\textbf{Trainable action decoding.}
The MVAE decoder was originally trained
to reconstruct actions from ground-truth UVTs $u_t$.
During VLA training, however,
the policy produces predicted UVTs $\hat{u}_t$
that may not exactly match the distribution
of precomputed $u_t$.
If the decoder were frozen,
small deviations in $\hat{u}_t$
could lead to amplified reconstruction errors.

We therefore keep the action decoder $g^a$ trainable
and apply an additional action decoding loss
\[
\hat{a}_t = g^{a}(\hat{u}_t),
\qquad
\mathcal{L}_{\text{dec}}
=
\| \hat{a}_t - a_t \|_1.
\]

This allows the decoder to adapt
to the distribution of predicted UVTs
while preserving the UVT alignment objective.
The UVT loss anchors the policy
in unified visuomotor target space,
and the decoding loss ensures that decoded outputs
remain consistent with executable robot actions.

The final objective is
\[
\mathcal{L}_{\text{VLA}}
=
\mathcal{L}_{\text{UVT-align}}
+
\gamma \mathcal{L}_{\text{dec}}.
\]

Note that this modification changes only the prediction target
and the loss formulation.
The backbone VLA architecture remains unchanged.
Any policy head capable of producing a UVT vector
$\hat{u}_t$ can be trained under this objective,
including regression-style heads (L2 on $\hat{u}_t$) and flow-matching
heads that generate $u_t$ in place of $a_t$ before action decoding.

\section{Experiments}
\label{sec:experiments}

\begin{table*}[!htbp]
\begin{center}
\caption{\textbf{Training Efficiency Comparison.} We compare our model's success rate under 10k and 100k training steps with baseline VLA-Adapter~\cite{adapter} across 2 different benchmarks, LIBERO~\cite{libero} and LIBERO-Plus~\cite{liberoplus}. We observed significant bigger model performance boost under limited computation at 10k steps, especially on the more challenging LIBERO-Plus benchmark, which introduces additional visual perturbations.}
\label{tab:libero_suite_comparison}
\begin{tabular}{l|l|cc|cc|cc|cc|cc}
\hline
& & \multicolumn{2}{c}{Object} & \multicolumn{2}{c}{Spatial} & \multicolumn{2}{c}{Goal} & \multicolumn{2}{c}{Long} & \multicolumn{2}{c}{Avg.} \\
\cline{3-4} \cline{5-6} \cline{7-8} \cline{9-10} \cline{11-12}
Environment & Method & 10k & 100k & 10k & 100k & 10k & 100k & 10k & 100k & 10k & 100k \\
\hline
\multirow{2}{*}{LIBERO~\cite{libero}} & VLA-Adapter~\cite{adapter} & 93.4 & 97.6 & 78.8 & 96.2 & 90.4 & 95.6 & 56.2 & 89.0 & 79.7 & 94.6 \\
 & \shortours~(Ours) & \textbf{94.0} & \textbf{99.6} & \textbf{95.0} & \textbf{98.6} & \textbf{95.0} & \textbf{97.2} & \textbf{68.8} & \textbf{91.2} & \textbf{88.2} & \textbf{96.7} \\
\hline
\multirow{2}{*}{LIBERO-Plus~\cite{liberoplus}} & VLA-Adapter~\cite{adapter} & \textbf{52.7} & 58.3 & 34.0 & 86.8 & 42.3 & 72.4 & 41.7 & 64.3 & 42.7 & 70.5 \\
 & \shortours~(Ours) & 52.4 & \textbf{60.4} & \textbf{81.5} & \textbf{87.2} & \textbf{67.8} & \textbf{74.7} & \textbf{63.0} & \textbf{72.1} & \textbf{66.2} & \textbf{73.6} \\
\hline
\end{tabular}
\end{center}
\vspace{-10 pt}
\end{table*}

\subsection{Experimental Setup}

We first describe the benchmarks, models, and training protocol used in our experiments.
We then present results on training efficiency (\cref{sec:training_efficiency}),
benchmark comparisons (\cref{sec:benchmark_results}),
real-world evaluation (\cref{sec:real_world}),
ablation studies (\cref{sec:ablation}),
and UVT representation analysis (\cref{exp:latent_analysis}).

\paragraph{Benchmarks.}
We evaluate our method primarily on the LIBERO benchmark suite~\cite{libero},
which consists of diverse manipulation tasks grouped into four categories:
\textit{Object}, \textit{Spatial}, \textit{Goal}, and \textit{Long} tasks.
Each suite poses distinct challenges in terms of complex tasks and environmental variation,
and success is measured by task success rate following the standard LIBERO protocol.

In addition, we evaluate on the LIBERO-Plus benchmark~\cite{liberoplus},
a recently released extension that introduces additional perturbations
and bigger visual and object state variations (e.g., perturbations in object configuration
and environment dynamics) to stress test model robustness.
LIBERO-Plus is designed to evaluate not only final performance
but also the stability and robustness of policies under more
challenging variations beyond the standard suites.

\paragraph{Models.}
To demonstrate the generality of our approach,
we integrate our \shortours~supervision framework into two
representative VLA systems with different backbones and
policy heads:
(1) \textbf{VLA-Adapter}~\cite{adapter}, which adopts a regression-style
action prediction head on top of a pretrained VLM backbone,
and (2) $\boldsymbol{\pi_{0.5}}$~\cite{pi}, which includes a flow-matching-based action expert with a distinct backbone configuration.
These models differ in both backbone architecture and
policy head design.
Our method modifies only the supervision target and
decoder integration described in~\cref{app:integration},
leaving the backbone and policy architecture otherwise unchanged.

\paragraph{Unified Visuomotor Target.}
The unified visuomotor target $u_t \in \mathbb{R}^{32}$
is obtained from the precision-aware MVAE described in \cref{sec:uvt_design}.
The MVAE is trained in a separate stage on paired robot actions $a_t$
and discrete dynamics codes $z_t$.
After training, we precompute $u_t$ for all training examples
and use them as targets during VLA fine-tuning.

\paragraph{Training Protocol.}
During fine-tuning, the VLA model predicts $\hat{u}_t$ from visual and
language input $x_t$, and a trainable MVAE decoder maps
$\hat{u}_t$ to a predicted action $\hat{a}_t$.
We optimize a combined objective
\[
\mathcal{L}_{\text{VLA}} =
\| \hat{u}_t - u_t \|_2^2
+
\gamma \| \hat{a}_t - a_t \|_1,
\]
where $\gamma=0.5$ balances UVT alignment with direct action reconstruction.
All models are trained under identical data splits, batch sizes,
and optimization schedules to ensure a fair comparison.
For training efficiency, we report success at fixed step budgets (10k and 100k).

\paragraph{Real-World Evaluation.}
For real-world evaluation, we design three challenging dual-arm manipulation tasks that test VLA policy’s ability to handle three distinct challenges: synchronization, precision, and cooperative manipulation.

\paragraph{Evaluation Metrics.}
We measure:
(1) success rate on LIBERO and LIBERO-Plus benchmarks,
(2) training efficiency (performance vs. training steps),
and (3) decoded action quality if applicable
(e.g., smoothness and UVT geometry analysis).

\subsection{Training Efficiency Analysis}
\label{sec:training_efficiency}
We first analyze training efficiency under fixed optimization budgets.
Table~\ref{tab:libero_suite_comparison} reports
success rates at fixed training steps
(10k and 100k).

\noindent\textbf{Early-stage performance gains.}
We observe improved training efficiency with \shortours, achieving higher success rates at early training checkpoints.
As seen in \cref{tab:libero_suite_comparison}, at 10k steps (around 8 GPU hours on A6000),
our method improves success rate
from 78.8\% to 95.0\%(+16.2\%) on Spatial tasks
and from 56.2\% to 68.8\%(+12.6\%) on Long tasks.
Similar gains are also observed on Object and Goal suites.
Importantly, these gains occur
without increasing model capacity
or modifying the backbone architecture
(Sec.~\ref{app:integration}).

\noindent\textbf{Saturation performance improvements.}
At longer training horizons (100k steps),
our method matches or slightly exceeds
the baseline across all suites.
For example, on the Object suite,
performance increases from 97.6\% to 99.6\%.
This indicates that UVT supervision
does not trade final accuracy for speed,
but instead improves optimization
while preserving asymptotic performance.

\noindent\textbf{Robustness under perturbation.}
The acceleration effect persists
under the more challenging LIBERO-Plus benchmark.
At 10k steps, our method substantially outperforms
the baseline on Spatial (81.5\% vs. 34.0\%)
and Long (63.0\% vs. 41.7\%) tasks.
Even under increased environmental perturbation,
UVT supervision stabilizes training
and improves early generalization.

Across all four suites and both benchmarks, reformulating imitation learning
as unified visuomotor target prediction accelerates VLA learning
without architectural modification or additional data.

\begin{table}[!htbp]
\caption{\textbf{LIBERO Benchmark Results.} We report the success rates of different VLA models across the four LIBERO benchmark suites, along with the overall average. Bold values denote the best performance.}
\label{tab:libero_benchmark}
\begin{center}
\begin{tabular}{l|cccc|c}
\hline
Model & Object & Spatial & Goal & Long & Avg. \\
\hline
VLA-Adapter~\cite{adapter} & 97.6 & 96.2 & 95.6 & 89.0 & 94.6 \\
$\pi_{0.5}$~\cite{pi} & 97.8 & 98.4 & \textbf{98.0} & 90.6 & 96.2 \\
\shortours~(VLA-Adapter) & \textbf{99.6} & \textbf{98.6} & 97.2 & 91.2 & 96.7 \\
\shortours~($\pi_{0.5}$) & 99.0 & 98.4 & 97.8 & \textbf{92.4} & \textbf{96.9} \\
\hline
\end{tabular}
\end{center}
\end{table}

\subsection{Benchmark Results}
\label{sec:benchmark_results}

Table~\ref{tab:libero_benchmark} reports the final success rates on the four LIBERO suites.
Integrating \shortours~consistently improves performance
over the corresponding baselines without increasing model size.
For VLA-Adapter, our method improves the overall average success rate from 94.6 to 96.7,
achieving the best performance on the Object and Spatial suites.
Similarly, when applied to $\pi_{0.5}$, our method increases the average from 96.2 to 96.9,
with gains on Object and Long tasks, showing that \shortours~is compatible with different types of existing VLA architectures and gives consistent improvements.
Rollout videos on the LIBERO suites are available on our \webpage.

\FloatBarrier

\begin{figure*}[!t]
\centering
\includegraphics[width=\textwidth]{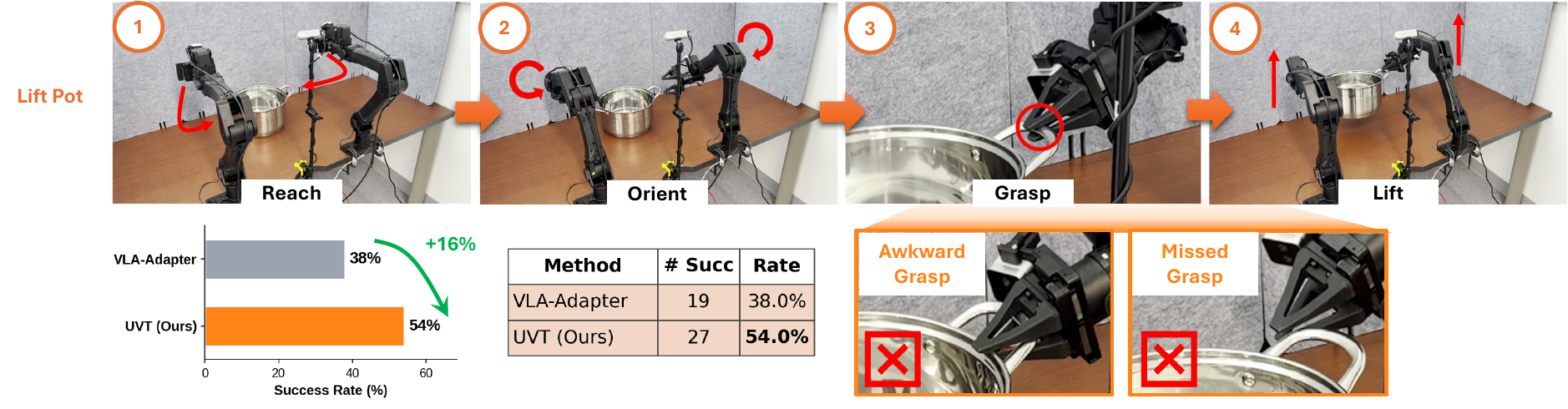}
\caption{\textbf{Lift Pot.}
Keyframes from a real-world rollout of our \shortours~policy: both arms \textit{Reach} toward the pot, \textit{Orient} their grippers toward the side handles, \textit{Grasp}, and \textit{Lift} the pot off the table.
The \textit{Grasp} stage is the main bottleneck: the zoom-ins show the baseline's dominant failure modes, an \textit{awkward grasp} closing on the rim or a \textit{missed grasp} missing the handles; since both handles must be secured simultaneously, either failure by either arm fails the trial. See videos on our \webpage.}
\label{fig:lift_pot}
\end{figure*}

\begin{figure*}[!t]
\centering
\includegraphics[width=\textwidth]{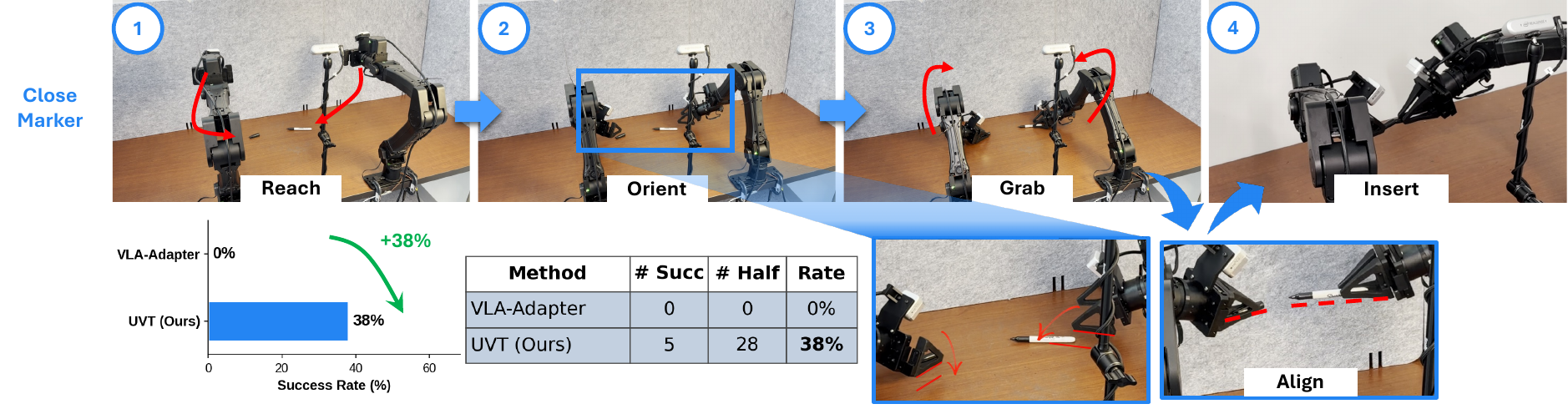}
\caption{\textbf{Close Marker.}
Keyframes from \shortours~policy: the arms \textit{Reach} toward the marker and cap, \textit{Orient} their grippers parallel to the table, \textit{Grab} the marker body and cap with separate grippers, then \textit{Align} and \textit{Insert} the cap onto the marker.
This is the most precision-demanding task: the zoom-in panels highlight the \textit{Align} steps, where sub-centimeter accuracy is required.
\# Half denotes half successes where grippers pick up both objects successfully but fail the insertion.}
\label{fig:close_marker}
\end{figure*}

\begin{figure*}[!t]
\centering
\includegraphics[width=\textwidth]{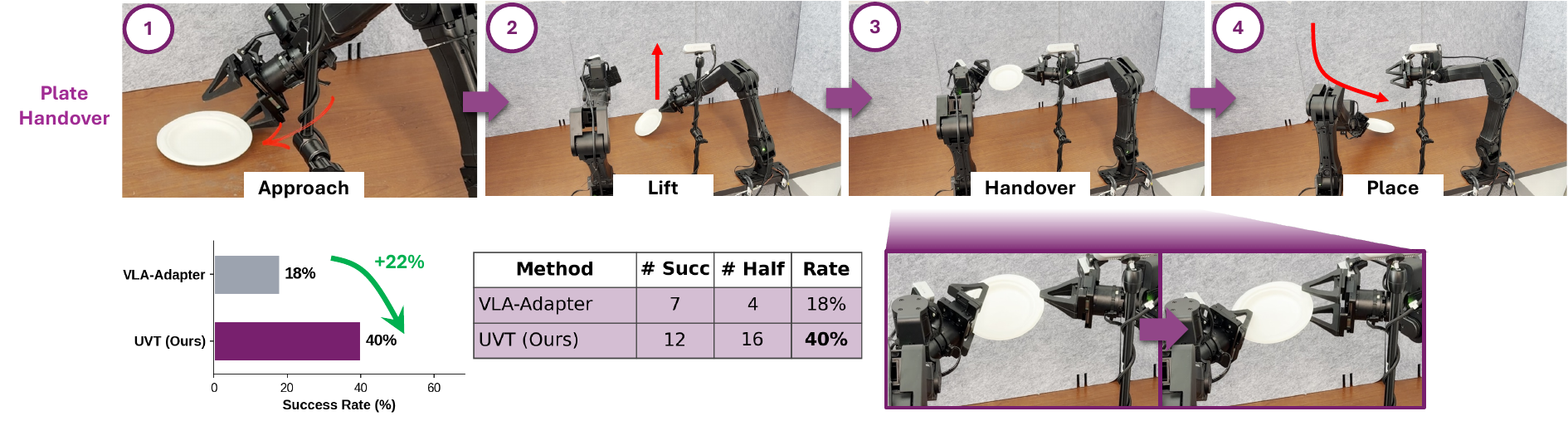}
\caption{\textbf{Plate Handover.}
Keyframes from \shortours~policy: the right arm \textit{Approaches} and slides below the thin paper plate, \textit{Lifts} it off the table, performs the dual-arm \textit{Handover}, and the left arm \textit{Places} it back down.
The most challenging step is the initial approach, where a gripper finger must slide underneath the thin plate, and the handover stage, where both grippers have to coordinate the movement. \# Half denotes half successes where gripper picks up the plate but fails the handover.}
\vspace{-10 pt}
\label{fig:plate_handover}
\end{figure*}

\subsection{Real-World Evaluation}
\label{sec:real_world}

We also evaluate \shortours~on a real bimanual manipulation platform under the same fine-tuning budget as the baseline (VLA-Adapter~\cite{adapter}).
We consider three tabletop tasks.
\textit{Lift Pot} requires the robot to grasp both side handles of a large aluminum pot and lift it into the air.
\textit{Close Marker} requires the robot to grasp a marker and its cap with separate grippers, align them, and close the marker.
\textit{Plate Handover} requires the robot to grasp a thin paper plate from the table with the right arm, transfer it to the left arm, and place it back on the table.
Example rollouts of our \shortours~policy on each task, together with per-task success rates and characteristic failure modes, are shown in \cref{fig:lift_pot,fig:close_marker,fig:plate_handover}; video illustrations of each task and its failure modes are available on our \webpage.
For each task, we collect 30 human demonstrations using VR headset teleoperation, fine-tune both methods for 10 GPU hours (A6000), and run 50 evaluation trials.
During both demonstration collection and evaluation, we apply the same amount of positional variation in initial object placement; for Lift Pot and Close Marker we additionally randomize the object rotation so the grippers are not always aligned with the object(s).

\noindent\textbf{Metric.}
For tasks where intermediate progress is meaningful, we record both full success and half-success and compute
\[
\text{Success Rate}=\frac{\#\text{Full}+0.5\times \#\text{Half}}{50}.
\]
Half-success for each task is defined as follows: successfully lifting the plate but failing the transfer for \textit{Plate Handover}, and successfully grasping both marker and cap but failing the insertion for \textit{Close Marker}.
\textit{Lift Pot} is evaluated with full success only.
All results are reported in \cref{fig:lift_pot,fig:close_marker,fig:plate_handover}.

\noindent\textbf{\shortours~improves object acquisition and secure grasps.}
Across tasks, we observe that \shortours~completes the initial grasping phases more reliably, which reduces downstream failures.
On \textit{Plate Handover} especially, \shortours~reaches the handover stage far more often: the baseline achieves only 4 half-successes and 7 full successes, whereas \shortours~achieves 16 half-successes and 12 full successes (\cref{fig:plate_handover}) (failure-mode rollouts can be viewed on our \webpage).
On \textit{Close Marker}, the baseline fails on every trial (0 full, 0 half), whereas \shortours~achieves 5 full successes and 28 half-successes: it grasps both the marker and cap consistently, and completes the precision insertion step on some trials.
On \textit{Lift Pot}, \shortours~improves lift success from 19/50 to 27/50 (38\%$\rightarrow$54\%). In many baseline failures, the handle slips shortly after lifting begins because the initial grasp is unstable or missed.

\noindent\textbf{Connection to decoded action behavior.}
We also observe that \shortours~policies tend to approach objects with fewer corrective micro-motions before contact, which reduces accidental grazing and improves grasp stability.
This is consistent with our offline analysis that decoded \shortours~actions are temporally smoother than direct action regression (Fig.~\ref{fig:smoothness}).
At the same time, \textit{Close Marker} highlights a remaining limitation even with our method: although \shortours~completes the full insertion on a handful of trials, the precise alignment of the marker to its cap still poses a significant challenge despite the substantially improved object acquisition (for more details please see representative trials on our \webpage).

\begin{table}[!htbp]
\caption{\textbf{Ablation on LIBERO-Spatial.} Comparison between (i) baseline action-only training, (ii) two-head auxiliary dynamics prediction without MVAE fusion using different LAM models, and (iii) \shortours~supervision. Results are reported at 10k and 100k training steps.}
\vspace{-5pt}
\label{tab:ablation}
\begin{center}
\begin{tabular}{lcc}
\hline
Method & @ 10k steps & @ 100k steps \\
\hline
VLA-Adapter~\cite{adapter} & 78.8 & 96.2 \\
w/ Villa-X~\cite{villax} head & 83.2 & 96.8 \\
w/ UniVLA~\cite{univla} head & 86.4 & 95.8 \\
\shortours~(Ours) & 95.0 & 98.6 \\
\hline
\end{tabular}
\end{center}
\vspace{-5pt}
\end{table}

\subsection{LAM Regularization Ablation}
\label{sec:ablation}
We study alternative ways of incorporating discrete dynamics codes
into VLA training to understand the role of shared UVT supervision.
Ablation results on different approaches on LIBERO-Spatial are summarized in Table~\ref{tab:ablation}.

\noindent\textbf{Two-head regularization.}
A natural first attempt to incorporate discrete dynamics codes is to duplicate the policy head
and add an auxiliary prediction task for discrete dynamics codes.
Under this design,
the VLA model predicts robot actions
through its original head
while an additional head predicts the discrete dynamics code $z_t$.
The auxiliary cross-entropy loss is expected
to regularize the shared VLM hidden states to encode more dynamics information.

We evaluate this strategy using two different Latent Action Models,
Villa-X~\cite{villax} and UniVLA~\cite{univla}.
As shown in Table~\ref{tab:ablation},
adding a second policy head does improve early performance
relative to the baseline.
At 10k steps, success increases from 78.8\%
to 83.2\% (Villa-X) and 86.4\% (UniVLA).
However, the gains are moderate,
and final performance at 100k steps
remains similar to the baseline.

\noindent\textbf{Unified visuomotor target.}
In contrast, our method constructs
a shared unified visuomotor target $u_t$
that fuses robot action and discrete dynamics information
into a single supervision space.
Instead of predicting actions and dynamics independently,
the policy aligns to a common UVT
in which both signals are jointly encoded.

This design substantially accelerates training without adding the training overhead of an additional policy head.
At 10k steps,
success improves to 95.0\%,
a large margin over both the baseline
and the two-head variants.
Final performance at 100k steps
also increases to 98.6\%.

\noindent\textbf{Discussion.}
The comparison shows that
simply adding an auxiliary dynamics prediction loss
is insufficient to fully exploit discrete dynamics information.
Independent heads supervise separate outputs,
but do not enforce interaction between motor
and dynamics representations.
By contrast, \shortours~encodes both signals jointly,
showing stronger regularization
and significantly faster convergence.

\subsection{Unified Visuomotor Target Analysis}
\label{exp:latent_analysis}

\begin{figure}[!t]
\centering
\includegraphics[width=\columnwidth]{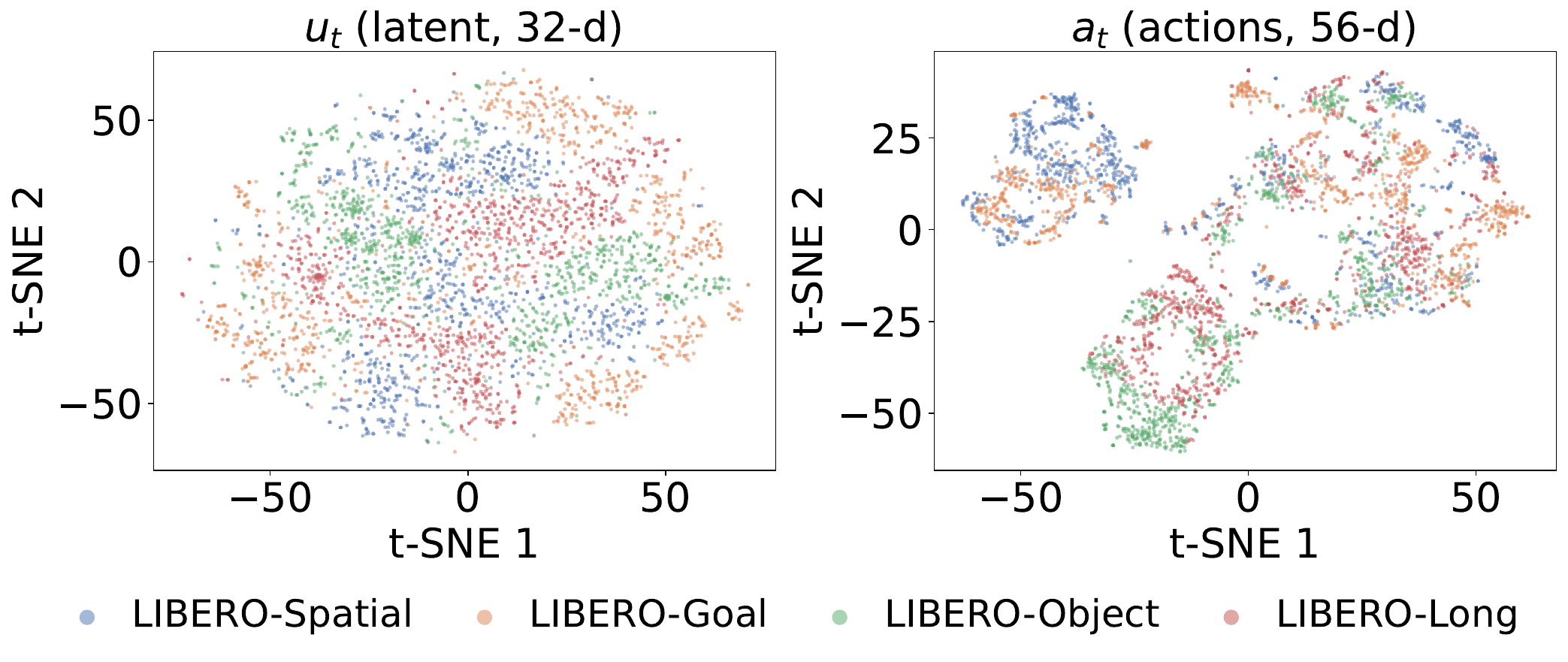}
\caption{\textbf{t-SNE visualization of unified visuomotor target $u_t$ and robot action $a_t$.}
Compared to robot actions, the \shortours~$u_t$ exhibits a more structured and smoothly organized embedding across LIBERO suites. We can observe that even without suite information, $u_t$ nicely groups actions from the same suite together, giving a more semantically consistent representation of task dynamics and motor behaviors.}
\vspace{-10pt}
\label{fig:tsne}
\end{figure}
We analyze the unified visuomotor target $u_t$ in comparison to robot actions $a_t$.
We first visualize both representations using t-SNE~\cite{tsne} over trajectories from the four LIBERO suites,
as shown in \cref{fig:tsne}. Although suite identifiers are never provided during MVAE training,
$u_t$ exhibits clear and smooth grouping by suite, whereas robot action embeddings $a_t$
appear heavily intermixed.

This difference reflects the type of information encoded by each representation.
Robot actions primarily contain low-level motor control signals.
Similar actuator-level patterns (e.g., small corrections, gripper adjustments)
frequently occur across different suites, even when the underlying task objectives differ.
As a result, action vectors from different suites often overlap numerically,
leading to entangled distributions in action space.
In contrast, $u_t$ is trained with additional supervision from the discrete dynamics code $z_t$,
which captures high-level visual dynamics across the entire interval $[t, t+k]$.
By incorporating this visual transition signal, $u_t$ emphasizes task-dependent motion patterns rather than embodiment-specific robot actions, hence the clearer grouping in the representation space.

The improved suite-level grouping is especially helpful for VLA prediction.
Different suites correspond to different task families and interaction dynamics.
If the supervision target clusters by suite without access to suite labels,
it indicates that the representation encodes high-level task-dependent information
that is consistently implied by the visual-language inputs.
Compared to robot actions $a_t$,
$u_t$ aligns more consistently with task context.
Such a target reduces irrelevant variability and makes the conditional mapping
from $x_t$ to the supervision signal more aligned with the underlying task structure,
which can help optimization.

\begin{figure}[!t]
\centering
\includegraphics[width=\columnwidth]{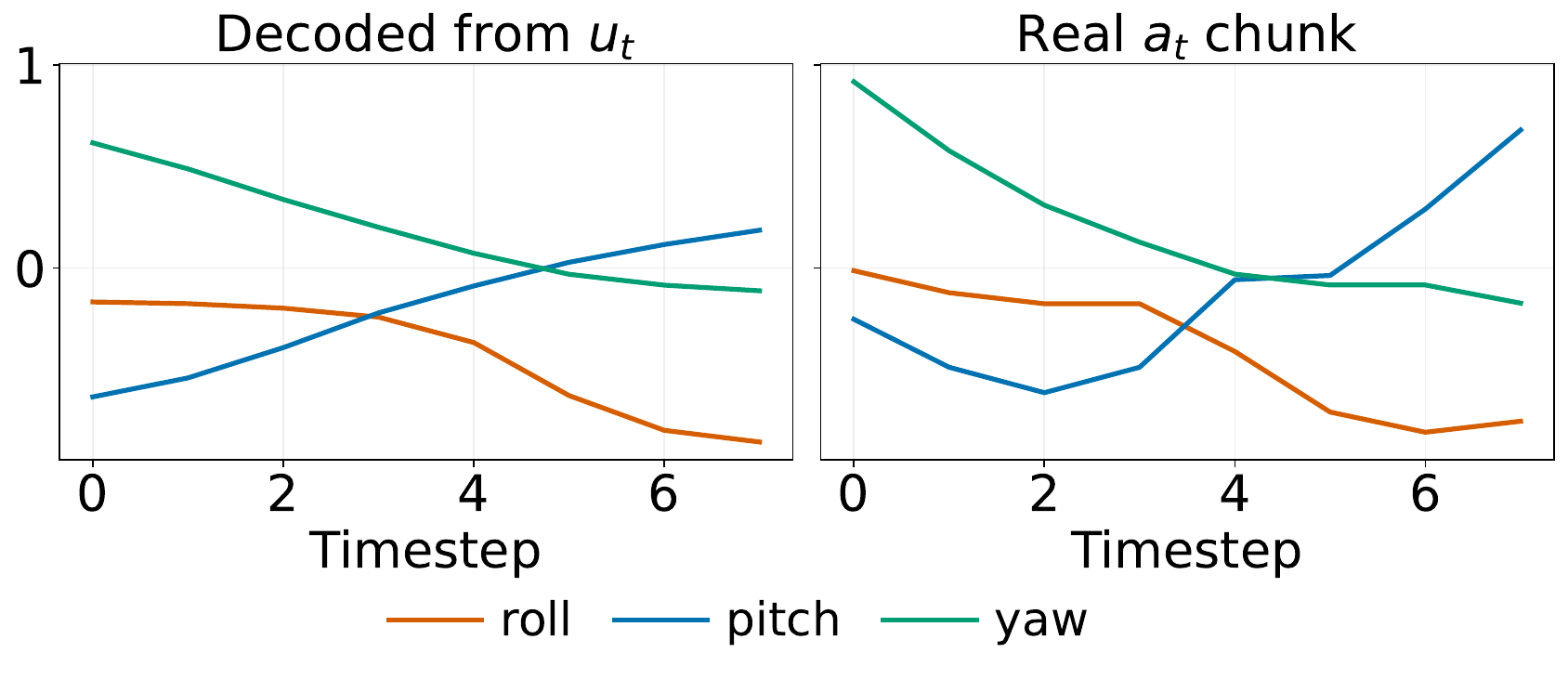}
\vspace{-15pt}
\caption{\textbf{Temporal profiles of decoded vs. demonstration action trajectories.}
Robot actions decoded from predicted $\hat{u}_t$ (left) follow the same movement trend as the corresponding demonstration action chunk $a_t$ (right) while exhibiting visibly fewer high-frequency fluctuations.}
\label{fig:smoothness}
\vspace{-15pt}
\end{figure}

Beyond the geometric structure of the UVT space, we analyze the temporal profiles of the decoded actions. We sample predicted $\hat{u}_t$ from the trained policy, decode them into robot actions, and compare the resulting trajectories with the original demonstration actions. As shown in \cref{fig:smoothness}, the decoded trajectories are noticeably smoother, exhibiting fewer high-frequency fluctuations while preserving the overall movement trend (corresponding rollouts on our \webpage), suggesting that the learned UVT space captures the underlying motion dynamics while filtering out low-level actuator noise present in demonstration robot action trajectories.

Together, the clearer suite grouping and smoother decoded trajectories
show that $u_t$ forms a supervision target that emphasizes visual dynamics information
and suppresses low-level robot action variability.
By shifting the prediction objective from purely low-level control signals
to a representation that also encodes high-level dynamics,
$u_t$ provides a better-conditioned target for VLA models,
which achieves improved training efficiency and overall VLA performance.

\textbf{Acknowledgments.} We thank the IROS reviewers and Area Chair for their insightful feedback, as well as Basavasagar Patil and Dwip Dalal for their careful review of our manuscript. We also thank Yuchen Song for assistance with the robotic setup and Aditya Mittal for discussions.

\section{Conclusion}
\label{sec:conclusion}

We presented \shortours, a supervision reformulation for VLA imitation learning that replaces direct action regression with unified visuomotor target prediction, combining robot actions with image-derived discrete dynamics signals. This change requires no modification to VLA backbones or policy architectures and applies broadly across VLA designs. \shortours~consistently accelerates fine-tuning and improves early success under fixed training budgets on LIBERO and LIBERO-Plus, while also yielding performance gains on real-world bimanual manipulation tasks.

Looking ahead, better prediction targets may also help beyond tabletop VLAs.
Language-driven navigation~\cite{shah2023vint,chen2019exploration,wasserman2022sling,wasserman2024xgx,puig2024habitat3}
similarly couples high-level scene understanding with low-level control;
multi-agent coordination~\cite{stone2010adhoc,rashid2018qmix,jain2019twobody,jain2020cordialsync,patel2021comon}
requires predicting joint behaviors under partial observability;
and classical imitation-learning failure modes, including compounding error and suboptimal cloning objectives~\cite{simchowitz2025pitfalls,eysenbach2022imitating,weihs2021advisor,jain2021gridtopix}, motivate supervision signals that are easier to learn than raw actions.

{\small
\bibliographystyle{IEEEtran}
\bibliography{main}

\begin{thebibliography}{10}
\providecommand{\url}[1]{#1}
\csname url@samestyle\endcsname
\providecommand{\newblock}{\relax}
\providecommand{\bibinfo}[2]{#2}
\providecommand{\BIBentrySTDinterwordspacing}{\spaceskip=0pt\relax}
\providecommand{\BIBentryALTinterwordstretchfactor}{4}
\providecommand{\BIBentryALTinterwordspacing}{\spaceskip=\fontdimen2\font plus
\BIBentryALTinterwordstretchfactor\fontdimen3\font minus
  \fontdimen4\font\relax}
\providecommand{\BIBforeignlanguage}[2]{{%
\expandafter\ifx\csname l@#1\endcsname\relax
\typeout{** WARNING: IEEEtran.bst: No hyphenation pattern has been}%
\typeout{** loaded for the language `#1'. Using the pattern for}%
\typeout{** the default language instead.}%
\else
\language=\csname l@#1\endcsname
\fi
#2}}
\providecommand{\BIBdecl}{\relax}
\BIBdecl

\bibitem{rt2}
A.~Brohan, N.~Brown, J.~Carbajal, Y.~Chebotar, X.~Chen \emph{et~al.}, ``Rt-2:
  Vision-language-action models transfer web knowledge to robotic control,'' in
  \emph{RSS}, 2023.

\bibitem{pi0fast}
K.~Pertsch, K.~Stachowicz, B.~Ichter, D.~Driess, S.~Nair, Q.~Vuong, O.~Mees,
  C.~Finn, and S.~Levine, ``Fast: Efficient action tokenization for
  vision-language-action models,'' in \emph{RSS}, 2025.

\bibitem{vqvla}
Y.~Wang, H.~Zhu, M.~Liu, J.~Yang, H.-S. Fang, and T.~He, ``Vq-vla: Improving
  vision-language-action models via scaling vector-quantized action
  tokenizers,'' in \emph{ICCV}, 2025.

\bibitem{openvla_oft}
M.~J. Kim, C.~Finn, and P.~Liang, ``Fine-tuning vision-language-action models:
  Optimizing speed and success,'' \emph{arXiv:2502.19645}, 2025.

\bibitem{pi0}
K.~Black, N.~Brown, D.~Driess, A.~Esmail, M.~Equi, C.~Finn, N.~Fusai, L.~Groom,
  K.~Hausman, B.~Ichter, S.~Jakubczak, T.~Jones, L.~Ke, S.~Levine, A.~Li-Bell,
  M.~Mothukuri, S.~Nair, K.~Pertsch, L.~X. Shi, J.~Tanner, Q.~Vuong,
  A.~Walling, H.~Wang, and U.~Zhilinsky, ``$\pi_0$: A vision-language-action
  flow model for general robot control,'' in \emph{RSS}, 2025.

\bibitem{pi}
{Physical Intelligence} \emph{et~al.}, ``$\pi_{0.5}$: a vision-language-action
  model with open-world generalization,'' 2025.

\bibitem{adapter}
Y.~Wang, P.~Ding, L.~Li, C.~Cui, Z.~Ge, X.~Tong, W.~Song, H.~Zhao, W.~Zhao,
  P.~Hou \emph{et~al.}, ``Vla-adapter: An effective paradigm for tiny-scale
  vision-language-action model,'' \emph{arXiv:2509.09372}, 2025.

\bibitem{ki}
D.~Driess, J.~T. Springenberg, B.~Ichter, L.~Yu, A.~Li-Bell, K.~Pertsch, A.~Z.
  Ren, H.~Walke, Q.~Vuong, L.~X. Shi, and S.~Levine, ``Knowledge insulating
  vision-language-action models: Train fast, run fast, generalize better,''
  \emph{arXiv:2505.23705}, 2025.

\bibitem{anchoralign}
D.~Dalal, S.~Patel, C.~Jain, J.~Kim, U.~Mishra, A.~Baratian, H.~Ha, H.~Ji,
  S.~Lazebnik, and U.~Jain, ``Generalizable {VLA} finetuning via representation
  anchoring and language-action alignment,'' \emph{arXiv:2607.13429}, 2026.

\bibitem{lapa}
S.~Ye, J.~Jang, B.~Jeon, S.~Joo, J.~Yang, B.~Peng, A.~Mandlekar, R.~Tan, Y.-W.
  Chao, Y.~Lin, L.~Liden, K.~Lee, J.~Gao, L.~Zettlemoyer, D.~Fox, and M.~Seo,
  ``Latent action pretraining from videos,'' in \emph{ICLR}, 2025.

\bibitem{univla}
Q.~Bu, Y.~Yang, J.~Cai, S.~Gao, G.~Ren, M.~Yao, P.~Luo, and H.~Li, ``Learning
  to act anywhere with task-centric latent actions,'' in \emph{RSS}, 2025.

\bibitem{villax}
X.~Chen, H.~Wei, P.~Zhang, Z.~Wang, S.~Li \emph{et~al.}, ``villa-x: Enhancing
  latent action modeling in vision-language-action models,''
  \emph{arXiv:2507.23682}, 2025.

\bibitem{mvplam}
J.~M. Lee, D.~Lee, S.~Ju, T.~Cho, J.~W. Koo, L.~Zhao, S.~Hong, and J.~Lee,
  ``Mvp-lam: Learning action-centric latent action via cross-viewpoint
  reconstruction,'' \emph{arXiv:2602.03668}, 2026.

\bibitem{routray2025vipra}
S.~Routray, H.~Pan, U.~Jain, S.~Bahl, and D.~Pathak, ``{ViPRA}: Video
  prediction for robot actions,'' \emph{arXiv:2511.07732}, 2025.

\bibitem{libero}
B.~Liu, Y.~Zhu, C.~Gao, Y.~Feng, Q.~Liu, Y.~Zhu, and P.~Stone, ``{LIBERO}:
  Benchmarking knowledge transfer for lifelong robot learning,'' in
  \emph{NeurIPS D\&B Track}, 2023.

\bibitem{liberoplus}
S.~Fei, S.~Wang, J.~Shi, Z.~Dai, J.~Cai, P.~Qian, L.~Ji, X.~He, S.~Zhang,
  Z.~Fei, J.~Fu, J.~Gong, and X.~Qiu, ``{LIBERO-Plus}: In-depth robustness
  analysis of vision-language-action models,'' \emph{arXiv:2510.13626}, 2025.

\bibitem{openvla}
M.~J. Kim, K.~Pertsch, S.~Karamcheti, T.~Xiao, A.~Balakrishna, S.~Nair,
  R.~Rafailov, E.~Foster, G.~Lam, P.~Sanketi, Q.~Vuong, T.~Kollar,
  B.~Burchfiel, R.~Tedrake, D.~Sadigh, S.~Levine, P.~Liang, and C.~Finn,
  ``Openvla: An open-source vision-language-action model,'' in \emph{CoRL},
  2024.

\bibitem{chi2023diffusionpolicy}
C.~Chi, S.~Feng, Y.~Du, Z.~Xu, E.~Cousineau, B.~Burchfiel, and S.~Song,
  ``Diffusion policy: Visuomotor policy learning via action diffusion,'' in
  \emph{RSS}, 2023.

\bibitem{roboflamingo}
X.~Li, M.~Liu, H.~Zhang, C.~Yu, J.~Xu, H.~Wu, C.~Cheang, Y.~Jing, W.~Zhang,
  H.~Liu, H.~Li, and T.~Kong, ``Vision-language foundation models as effective
  robot imitators,'' in \emph{ICLR}, 2024.

\bibitem{oxe}
A.~Padalkar, A.~Pooley, A.~Jain, A.~Bewley, A.~Herzog, A.~Irpan, A.~Khazatsky,
  A.~Rai, A.~Singh, A.~Brohan \emph{et~al.}, ``Open x-embodiment: Robotic
  learning datasets and rt-x models,'' in \emph{CoRL}, 2023.

\bibitem{octo}
{Octo Model Team} \emph{et~al.}, ``Octo: An open-source generalist robot
  policy,'' in \emph{RSS}, 2024.

\bibitem{bahl2023vrb}
S.~Bahl, R.~Mendonca, L.~Chen, U.~Jain, and D.~Pathak, ``{Affordances from
  Human Videos as a Versatile Representation for Robotics},'' in \emph{CVPR},
  2023.

\bibitem{dasari2023data4robotics}
S.~Dasari, M.~K. Srirama, U.~Jain, and A.~Gupta, ``{An Unbiased Look at
  Datasets for Visuo-Motor Pre-Training},'' in \emph{CoRL}, 2023.

\bibitem{molmoact}
J.~Lee, J.~Duan, H.~Fang, Y.~Deng, S.~Liu, B.~Li, B.~Fang, J.~Zhang, Y.~R.
  Wang, S.~Lee, W.~Han, W.~Pumacay, A.~Wu, R.~Hendrix, K.~Farley,
  E.~VanderBilt, A.~Farhadi, D.~Fox, and R.~Krishna, ``{MolmoAct}: Action
  reasoning models that can reason in space,'' \emph{arXiv:2508.07917}, 2025.

\bibitem{patel2026rigvid}
S.~Patel, S.~Mohan, H.~Mai, U.~Jain, S.~Lazebnik, and Y.~Li, ``Robotic
  manipulation by imitating generated videos without physical demonstrations,''
  in \emph{ICLR}, 2026.

\bibitem{genie}
J.~Bruce, M.~Dennis, A.~Edwards, J.~Parker-Holder, Y.~Shi, E.~Hughes, M.~Lai,
  A.~Mavalankar, R.~Steigerwald, C.~Apps, Y.~Aytar, S.~Bechtle, F.~Behbahani,
  S.~Chan, N.~Heess, L.~Gonzalez, S.~Osindero, S.~Ozair, S.~Reed, J.~Zhang,
  K.~Zolna, J.~Clune, N.~de~Freitas, S.~Singh, and T.~Rockt{\"a}schel, ``Genie:
  Generative interactive environments,'' in \emph{ICML}, 2024.

\bibitem{conla}
W.~Dai, K.~Lan, J.~Zhou, B.~Zhao, X.~Su, J.~Tong, W.~Guan, and S.~Yang,
  ``Conla: Contrastive latent action learning from human videos for robotic
  manipulation,'' \emph{arXiv:2602.00557}, 2026.

\bibitem{hinton2002poe}
G.~E. Hinton, ``Training products of experts by minimizing contrastive
  divergence,'' \emph{Neural Computation}, 2002.

\bibitem{zhao2023act}
T.~Z. Zhao, V.~Kumar, S.~Levine, and C.~Finn, ``Learning fine-grained bimanual
  manipulation with low-cost hardware,'' in \emph{RSS}, 2023.

\bibitem{wu2018multimodal}
M.~Wu and N.~Goodman, ``Multimodal generative models for scalable
  weakly-supervised learning,'' in \emph{NeurIPS}, 2018.

\bibitem{tsne}
L.~van~der Maaten and G.~Hinton, ``Visualizing data using {t-SNE},''
  \emph{JMLR}, 2008.

\bibitem{shah2023vint}
D.~Shah, A.~Sridhar, N.~Dashora, K.~Stachowicz, K.~Black, N.~Hirose, and
  S.~Levine, ``{ViNT}: A foundation model for visual navigation,'' in
  \emph{CoRL}, 2023.

\bibitem{chen2019exploration}
T.~Chen, S.~Gupta, and A.~Gupta, ``Learning exploration policies for
  navigation,'' in \emph{ICLR}, 2019.

\bibitem{wasserman2022sling}
J.~Wasserman, K.~Yadav, G.~Chowdhary, A.~Gupta, and U.~Jain, ``Last-mile
  embodied visual navigation,'' in \emph{CoRL}, 2022.

\bibitem{wasserman2024xgx}
J.~Wasserman, G.~Chowdhary, A.~Gupta, and U.~Jain, ``Exploitation-guided
  exploration for semantic embodied navigation,'' in \emph{ICRA}, 2024.

\bibitem{puig2024habitat3}
X.~Puig, E.~Undersander, A.~Szot, M.~Cote, R.~Partsey, J.~Yang, R.~Desai,
  A.~Clegg, M.~Hlavac, T.~Min, T.~Gervet, V.~Vondrus, V.-P. Berges, J.~Turner,
  O.~Maksymets, Z.~Kira, M.~Kalakrishnan, J.~Malik, D.~Chaplot, U.~Jain,
  D.~Batra, A.~Rai, and R.~Mottaghi, ``{Habitat 3.0}: A co-habitat for humans,
  avatars and robots,'' in \emph{ICLR}, 2024.

\bibitem{stone2010adhoc}
P.~Stone, G.~A. Kaminka, S.~Kraus, and J.~S. Rosenschein, ``Ad hoc autonomous
  agent teams: Collaboration without pre-coordination,'' in \emph{AAAI}, 2010.

\bibitem{rashid2018qmix}
T.~Rashid, M.~Samvelyan, C.~Schroeder, G.~Farquhar, J.~Foerster, and
  S.~Whiteson, ``{QMIX}: Monotonic value function factorisation for deep
  multi-agent reinforcement learning,'' in \emph{ICML}, 2018.

\bibitem{jain2019twobody}
U.~Jain, L.~Weihs, E.~Kolve, M.~Rastegari, S.~Lazebnik, A.~Farhadi, A.~Schwing,
  and A.~Kembhavi, ``Two body problem: Collaborative visual task completion,''
  in \emph{CVPR}, 2019.

\bibitem{jain2020cordialsync}
U.~Jain, L.~Weihs, E.~Kolve, A.~Farhadi, S.~Lazebnik, A.~Kembhavi, and
  A.~Schwing, ``A cordial sync: Going beyond marginal policies for multi-agent
  embodied tasks,'' in \emph{ECCV}, 2020.

\bibitem{patel2021comon}
S.~Patel, S.~Wani, U.~Jain, A.~Schwing, S.~Lazebnik, M.~Savva, and A.~X. Chang,
  ``Interpretation of emergent communication in heterogeneous collaborative
  embodied agents,'' in \emph{ICCV}, 2021.

\bibitem{simchowitz2025pitfalls}
M.~Simchowitz, D.~Pfrommer, and A.~Jadbabaie, ``The pitfalls of imitation
  learning when actions are continuous,'' in \emph{COLT}, 2025.

\bibitem{eysenbach2022imitating}
B.~Eysenbach, S.~Udatha, S.~Levine, and R.~Salakhutdinov, ``Imitating past
  successes can be very suboptimal,'' in \emph{NeurIPS}, 2022.

\bibitem{weihs2021advisor}
L.~Weihs, U.~Jain, I.-J. Liu, J.~Salvador, S.~Lazebnik, A.~Kembhavi, and
  A.~Schwing, ``Bridging the imitation gap by adaptive insubordination,'' in
  \emph{NeurIPS}, 2021.

\bibitem{jain2021gridtopix}
U.~Jain, I.-J. Liu, S.~Lazebnik, A.~Kembhavi, L.~Weihs, and A.~Schwing,
  ``{GridToPix}: Training embodied agents with minimal supervision,'' in
  \emph{ICCV}, 2021.

\end{thebibliography}
}

\end{document}